%% file: main.tex
\documentclass[a4paper,10pt,conference]{IEEEtran}
\input{macros.tex}

\usepackage{graphicx}
\usepackage{amsmath}
\usepackage{amssymb}
\usepackage{tabularx}
\usepackage{amsmath}
\usepackage{makecell}
\usepackage{algorithmic}
\usepackage{amsmath}
\usepackage{mathtools}
\usepackage{graphics}
\usepackage{midfloat}
\usepackage{listings}
\usepackage[ruled]{algorithm2e}
\usepackage[utf8]{inputenc}
\usepackage{float}
\usepackage{multirow}
\usepackage{capt-of}
\usepackage{kantlipsum}
\usepackage{url}
\usepackage{paralist}
\usepackage{subcaption}
\usepackage{orcidlink}
\usepackage[left=1.57cm,right=1.57cm,top=0.95cm,bottom=2.54cm]{geometry}

\graphicspath{{img/}}

\begin{document}
\title{FedSAUC: A Similarity-Aware Update Control for Communication-Efficient Federated Learning in Edge Computing
}

\author{
    \IEEEauthorblockN{
        Ming-Lun~Lee\IEEEauthorrefmark{1}, Han-Chang~Chou\IEEEauthorrefmark{1}, and Yan-Ann~Chen\orcidlink{0000-0002-3348-6022}\IEEEauthorrefmark{1}\IEEEauthorrefmark{2}
    }
    \IEEEauthorblockA{
    \IEEEauthorrefmark{1}Dept. of Computer Science and Engineering, Yuan Ze University, Taoyuan, Taiwan\\
    \IEEEauthorrefmark{2}Innovation Center for Big Data and Digital Convergence, Yuan Ze University, Taoyuan, Taiwan\\
    E-mail: s1071417@mail.yzu.edu.tw, s1071443@mail.yzu.edu.tw, chenya@saturn.yzu.edu.tw}
}

\IEEEoverridecommandlockouts
\IEEEpubid{\makebox[\columnwidth]{(C)2021 IPSJ\hfill}
\hspace{\columnsep}\makebox[\columnwidth]{ }}

\maketitle
\IEEEpubidadjcol

\begin{abstract}
Federated learning is a distributed machine learning framework
to collaboratively train a global model without
uploading privacy-sensitive data onto a centralized server.
Usually, this framework is applied to
edge devices such as smartphones, wearable devices, and \emph{Internet of Things (IoT)} devices
which closely collect information from users.
However, these devices are mostly battery-powered.
The update procedure of federated learning
will constantly consume the battery power and
the transmission bandwidth.
In this work, we propose an update control for federated learning,
FedSAUC, by considering the similarity of users' behaviors (models).
At the server side,
we exploit clustering algorithms to
group devices with similar models.
Then we select some representatives for each cluster
to update information to train the model.
We also implemented a testbed prototyping on edge devices
for validating the performance.
The experimental results show that
this update control will not affect the training accuracy
in the long run.
\end{abstract}

\IEEEpeerreviewmaketitle

\section{Introduction} \label{Sec:Intro}
\emph{Federated learning {(FL)}} draws considerable attention
for its capability of training a general model from huge numbers of participants.
Traditional machine learning and deep learning algorithms requires
high-power computations, such as GPUs, to optimize parameters of training models.
For gaining more powerful computations,
we may further rent high performance computing units from cloud service providers.
This kind of computation paradigm requires a centralized computing server.
However, with the rapid growth of \emph{Internet of Things (IoT)},
billions of edge and wearable devices are ubiquitous in the world and
they will generate tons of data every day.
It would be impractical to upload such huge amount of data onto a centralized server
for storing and further processing
while considering the limited data storage, transmission bandwidth, and communication costs.
Moreover, users of wearable devices may not be willing to
upload their sensitive data, such as biological information, because of privacy concerns.
Google's researchers proposed federated learning \cite{mcmahan2017communication},
a distributed learning paradigm, to solve these problems.
\begin{figure}[!t]
\centering
\includegraphics[width=0.9\linewidth]{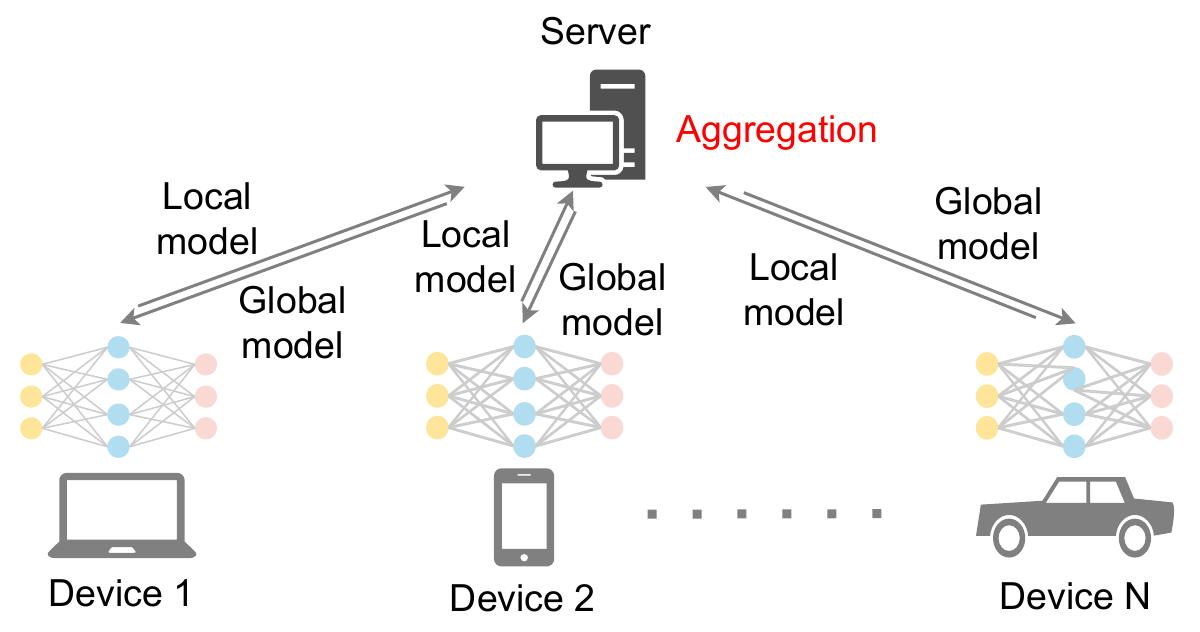}
\caption{The scenario of federated learning.}
\label{Fig:FL_scenario}
\end{figure}

\Fig{Fig:FL_scenario} shows the scenario of federated learning
where each device collaboratively train a global model
without sharing its own data.
These devices may be smartphones, smart watches, or edge devices.
Firstly, each device trains a local model with its collected data
by using an adopted learning algorithm.
Without directly uploading their personal data,
they upload parameters of their current local model instead.
Secondly, a global server will collect uploaded information from each device
and then aggregate all the information to form a global model.
Next, each device will receive the aggregated global model from the server.
Finally, this exchange procedure can repeat iteratively until the global model is converged.
However, federated learning still faces some challenges \cite{fl_chanllenges}:
expensive communication, systems heterogeneity, statistical heterogeneity, and privacy concerns.
Recent research reduces the communication cost between the server and a client
through model compression techniques and device sampling methods, but the similarity of users' behaviors is barely considered.
Furthermore, edge devices, e.g., smartphone and Raspberry Pi,
have limited computation power and energy constraints,
so we should also consider an energy-aware training mechanism
to enhance the system lifetime.

In this work, we investigate an update control mechanism
to suppress training tasks of some devices according to
the model similarity among devices.
For instance, if two devices sense similar condition or perform similar actions,
their local models may also be alike
since they are training the model with similar data distribution.
Here, we consider selecting partial devices
within a group of high similarity for a period of time
because the provided diversity to the global model may be close.
The suppressed devices may earn energy saving
for no training computation and update.
It also reduces messages received by the server.
Moreover, we evaluate the feasibility of
executing our proposed mechanism on edge devices.
The contributions of this work are as follows.
\begin{itemize}
\item We propose \emph{Similarity-Aware Update Control for Federated Learning (FedSAUC)}
to suppress training updates of some devices
where their local models are close to other devices.
That is to say that
active updating devices are representatives of a device group
with similar behaviors or collected information.
\item We evaluate the performance of several clustering algorithms
on similarity computations and
validate the training accuracy with and without our update suppression mechanism.
\item We build up a testbed for federated learning
by using off-the-shelf edge devices.
Thus, we can observe the feasibility of our approach
when running in the real-life environment.
\end{itemize}

The remainder of this paper is organized as follows.
We survey some related work on federated learning in \Sec{Sec:Related}.
\Sec{Sec:proposed} presents our detailed designs of FedSAUC.
We validate the performance of our approach by the prototyping result in \Sec{Sec:Experimental Results}.
\Sec{Sec:conc} concludes this paper.

\section{Related Work} \label{Sec:Related}
In recent years,
communication efficiency and power management are two major issues
which still remain open questions
in communications and computing research fields
and drive many researchers into the fields.
As the large amount of the devices
participating in a federated learning process,
the traffic between clients and server could be considerable
since it requires the continuous exchange of model information.
Moreover, federated learning used to
build upon IoT devices
which may have the constraint of energy consumption
such that the devices could run out of battery
and thus leave the training process,
resulting in a loss of information.
To solve these issues mentioned above,
researchers have proposed
decentralized federated learning \cite{hudecentralized},
federated learning with convergence preserve \cite{decentralized},
and federated learning with asynchronous training strategies \cite{semiasyn}, etc.
Some approaches \cite{dgc,expanding,qgmu}
deal with the amount of model parameters transferred
such as model compression by transmitting fraction of gradients.
Some of them also conduct client selection
which samples partial of devices
for improving the communication efficiency
and energy conservation during the training process.

%
For the purpose of saving the expensive communication cost,
traditional methods such as static sampling \cite{mcmahan2017communication}
samples partial of devices rather than every training device,
and random masking randomly selects fraction of model parameters
uploaded to the server.
However, static sampling exploits a fixed sample rate
during the training process
no matter what the model's states are.
Random sampling leads to some information loss
due to the partial selection of parameters.
Thus, reference \cite{ds} proposes
a dynamic sampling and selective masking mechanism.
They use dynamic sample rates by considering
the degree of decay with regard to the training epochs.
This mechanism samples more devices at the beginning of the training
to accelerate model convergence and
gradually decreases the device number.
On the other hand,
the selective masking measures the contribution
as the difference of each model parameter
between two consecutive training rounds during the training process.
Model parameters with the top-$k$ contributions
are selected for uploading to the server.
However, these methods have the potential risk
of losing partial statistical information
of the computed gradients.
Similarly, either static or dynamic sampling on devices
tends to be non-informed sampling
such that it selects devices based on the sample rate
but regardless of the training behavior of the devices.
In real-world federated learning,
devices may train on non-I.I.D. (non independent and identically distributed) data,
it is more important to select devices
that could form a coverage of all training labels
rather than the amount of training devices.
By considering the behaviors of devices' owners \cite{untargeted,gan},
we investigate the possibility of grouping devices according
to their behaviors (models) similarity.

\section{FedSAUC} \label{Sec:proposed}
\begin{figure}[!t]
\centering
\begin{tabular}{cc}
\includegraphics[width=0.43\linewidth]{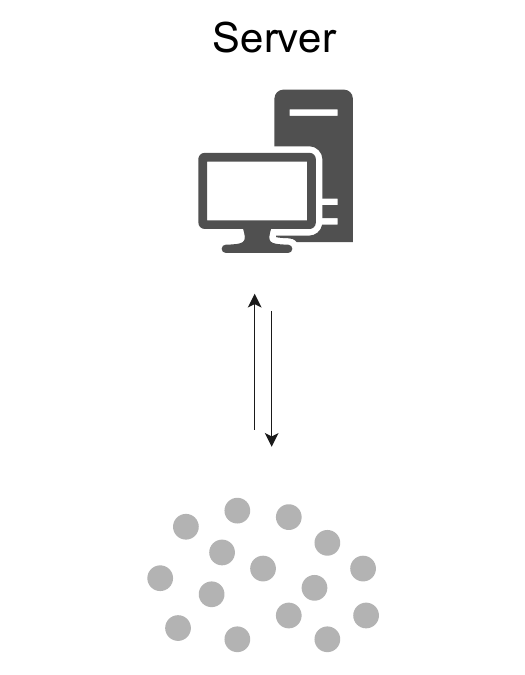}&
\includegraphics[width=0.43\linewidth]{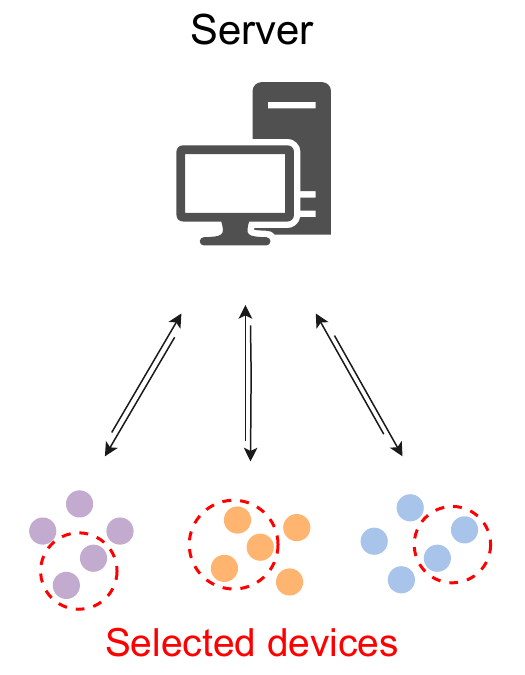}\\
(a)&(b)\\
\end{tabular}
\caption{Scenarios of (a) traditional federated learning
and (b) our proposed FedSAUC.}
\label{Fig:scenario}
\end{figure}

In this section, we address the problem formulation of our objective
and also introduce the flow of our proposed approach.
\Fig{Fig:scenario}(a) presents the architecture of traditional federated learning.
Each device contributes its local model onto a parameter server and
then it broadcasts a global model which is the aggregation of all the local models
to all devices.
We call this architecture a Parameter Server Architecture. 
For remedying the problem we addressed in \Sec{Sec:Intro},
we propose the update control mechanism which consists of
the analysis of model similarity and a update control method.
\Fig{Fig:scenario}(b) shows the concept of FedSAUC.
We form the devices which participate the learning process
into groups with similar models.
Then only partial devices are selected to update their local models
onto the server during a period of time.
\begin{figure}[!t]
\centering
\includegraphics[width=\linewidth]{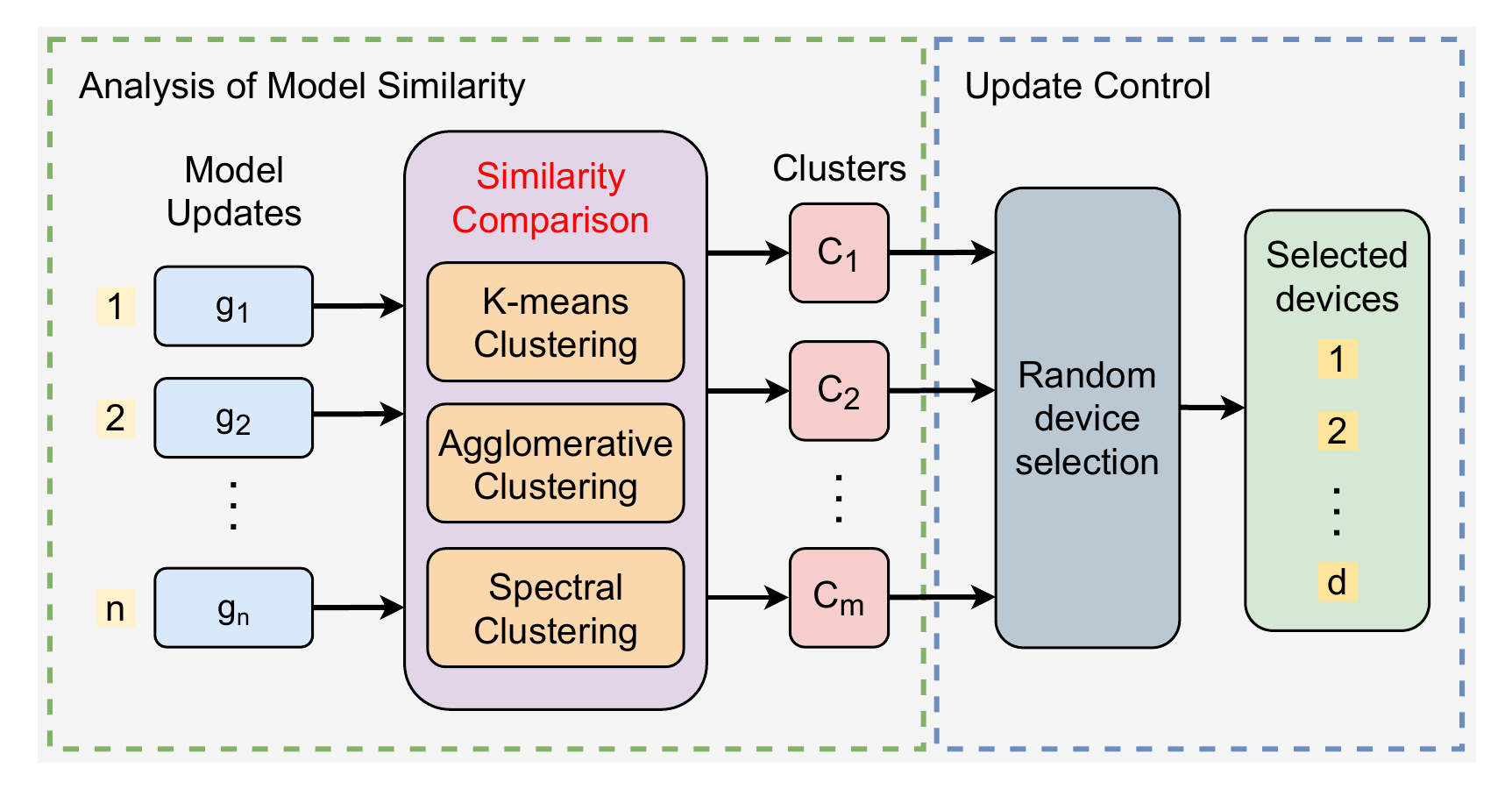}
\caption{The system architecture of FedSAUC.}
\label{Fig:flowChart}
\end{figure}
\Fig{Fig:flowChart} shows our system architecture
to realize the scenario.
We first utilize ``analysis of model similarity''
to form the device clusters and
then exploit ``update control''
to select the devices which have to update their model
to the server.

\subsection{Problem Formulation}
Assume that there are $n$ devices (clients)
which collaboratively train a machine learning model
without uploading their dataset $\{\mathcal{D}_i| i = 1..n\}$.
Let $\mathcal{L}(.)$ be the loss function of a global model
and $\mathcal{L}_i(.,.)$ be the loss function of device $i$.
We denote the model of a machine learning algorithm
by $\theta$, which represents parameters of an objective function.
The objective of traditional federated learning is to
find out $\theta$ such that the entire loss is minimized.
We formulate it as follows.
\begin{align}
\mathop{\min}_{\theta} \mathcal{L}(\theta) = \sum_{i=1}^n p_i \mathcal{L}_i(\theta,\mathcal{D}_i) \mathrm{,} \notag
\end{align}
where $p_i$ is a non-negative weight for device $i$'s loss.
$p_i$ is a hyper-parameter of the model
and here we set $p_i = {|\mathcal{D}_i|}/{\sum_{i=1}^{n}|D_i|}$,
the ratio of
the number of local data to that of all devices' data.
For obtaining the loss of device $i$,
we input its local data $\mathcal{D}_i$ into
the loss function $\mathcal{L}_i(. , .)$.
\begin{align}
\mathcal{L}_i(\theta, \mathcal{D}_i) = \frac{1}{|\mathcal{D}_i|} \sum_{\delta=1}^{|\mathcal{D}_i|}\mathcal{L}_i^\delta(\theta, \delta), \notag
\end{align}
where $\mathcal{L}_i^\delta(\theta, \delta)$ is the loss
with respect to an item $\delta$ of device $i$'s local dataset.

At the beginning of the training process,
each device shares a global model
which is from a central server.
Each device iteratively optimizes the shared model
with respect to its local dataset $D_i$.
For each round, each device will derive
a local model (weights) $\theta_i$ and
gradient updates for parameters $g_i$ from
the computation of optimizing the local loss.
Then the centralized server will aggregate
local models $\{\theta_i | i = 1..n\}$ from all the devices
and compute a global model $\theta$.
Next, the server broadcasts the new global model $\theta$
to all the devices.
Each device will repeat the training process
until the global model is converged.
However, it requires a large amount of transmissions
between the server and clients
and also results in large amounts of power consumption.

In FedSAUC, we address the problem of selecting
$d = |\kappa|$ devices such that $\mathcal{L}^{\kappa}(\theta^\kappa)$
is close to $\mathcal{L}(\theta)$
where $\kappa$ is the set of selected devices and $d << n$.
\begin{align}
\mathcal{L}^\kappa(\theta^\kappa) = \sum_{j=1}^{|\kappa|} p_j \mathcal{L}_j(\theta^\kappa,\mathcal{D}_j) \mathrm{,} \notag
\end{align}
where $p_j = {|\mathcal{D}_j|}/{\sum_{j=1}^{d}|D_j|}$.

\subsection{Analysis of Model Similarity} \label{Sec:msim}
During the update phase of traditional federated learning,
each device will send its gradient updates of parameters
to the server.
At the server side, it will collect gradient updates from all the devices,
$G = \{g_1, g_2, \dots, g_n\}$.
Intuitively, if two devices have similar datasets,
they may compute similar gradient updates
by using the same optimization process
and contributes duplicated information to the global model.
Therefore, in this component,
we utilize clustering algorithms for grouping
devices with similar gradient updates.
Here, we assume that we form $m$ device clusters
, $C = \{C_1, C_2, \dots, C_m\}$.


\textbf{K-means Clustering.}
K-means \cite{kmeans} is a basic clustering algorithm
which partitions $n$ data points,
$G = \{g_1, g_2, \dots, g_n\}$, into $m$ clusters.
The goal of K-means clustering is to minimize
\textit{WCSS (within-cluster sum of squares)}
which is the sum of distances from each data point
to its cluster centroid $\mu$:
\begin{equation}
\mathop{\arg\min}_{C} \sum_{k=1}^m\sum_{g \in {C_k}} \|{g - \mu_k} \|^2 \mathrm{,} \notag
\end{equation}
where $\mu_k$ is the centroid of points in cluster $C_k$.

\begin{figure}[t!]
\centering
\includegraphics[width=0.85\linewidth]{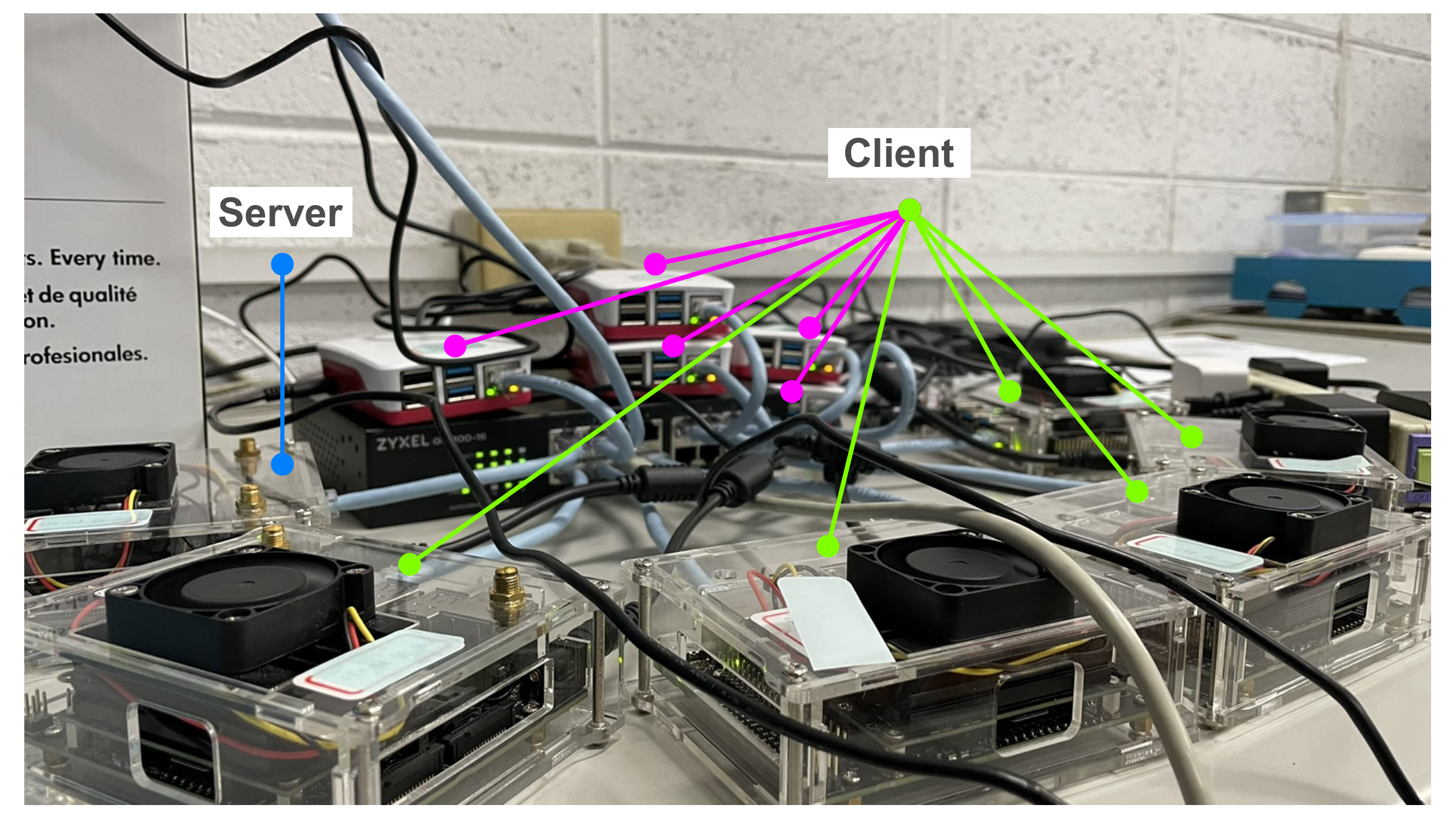}
\caption{Testbed prototyping: purple, green, and blue dots indicate Raspberry Pi 4, NVIDIA Jetson Nano (clients), and NVIDIA Jetson Nano (server), respectively.}
\label{Fig:Environment}
\end{figure}
\begin{table}[!t]
    \centering
    \captionof{table}{Device training configuration.}
\begin{tabular}{|l|c|c|c|c|c|}
    \hline
        Device & $v_0, v_1$ & $v_2, v_3$ & $v_4, v_5$ & $v_6, v_7$ & $v_8, v_9$  \\
    \hline
       Training Set 1 & 0, 1 & 2, 3 & 4, 5 & 6, 7 & 8, 9 \\
    \hline
       Training Set 2 & 0 & 2 & 4 & 6 & 8 \\
    \hline
\end{tabular}
\label{tab:device}
\end{table}
\begin{table}[t!]
    \centering
    \captionof{table}{Clustering accuracy.}
\begin{tabular}{|l|c|c|c|}
        \multicolumn{4}{c}{Training Set 1} \\
    \hline
        \multirow{2}{*}{Update types} &
        \multicolumn{3}{c|}{Clustering algorithms} \\
    \cline{2-4}
        & K-Means & Agglomerative & Spectral \\
    \hline
        WW & 62.6\% & 63.8\% & 78.6\% \\
    \hline
        WBW & 62.8\% & 63.8\% & 78.5\% \\
    \hline
        WG & 72.6\% & 74.6\% & 83.3\% \\
    \hline
        WBG & 72.9\% & 74.6\% & 83.7\% \\
    \hline
        \multicolumn{4}{c}{} \\
        \multicolumn{4}{c}{Training Set 2} \\
    \hline
        \multirow{2}{*}{Update types} &
        \multicolumn{3}{c|}{Clustering algorithms} \\
    \cline{2-4}
        & K-Means & Agglomerative & Spectral \\
    \hline
        WW & 99.8\% & 99.8\% & 100.0\% \\
    \hline
        WBW & 99.8\% & 99.8\% & 100.0\% \\
    \hline
        WG & 99.4\% & 99.4\% & 99.4\% \\
    \hline
        WBG & 99.4\% & 99.4\% & 99.4\% \\
    \hline
\end{tabular}
\label{tab:clustering}
\end{table}
\begin{table}[!t]
    \centering
    \captionof{table}{Message reductions after applying update control.}
\begin{tabular}{|l|c|c|c|c|}
        \multicolumn{5}{c}{Server side} \\
    \hline
        Starting rounds & round 10 & round 20 & round 30 & round 40 \\
    \hline
        Model parameters & 784K & 588K & 392K & 196K \\
    \hline
        Size of message & 17.50MB & 13.13MB & 8.75MB & 4.38MB \\
    \hline
        \multicolumn{5}{c}{} \\
        \multicolumn{5}{c}{Client side} \\
    \hline
        Starting rounds & round 10 & round 20 & round 30 & round 40 \\
    \hline
        Model parameters & 156.8K & 117.6K & 78.4K & 39.2K \\
    \hline
        Size of message & 3.50MB & 2.63MB & 1.75MB & 0.88MB \\
    \hline
\end{tabular}
\label{tab:reduction}
\end{table}

\textbf{Agglomerative Clustering.}
Agglomerative clustering \cite{agg}
is a hierarchical clustering method by
merging cluster pairs repeatedly
until there is only one cluster
where each cluster may contain a data point or
a group of data points.
In the merging step,
it looks for the minimal distance
among all the cluster pairs.
The distance measurement of a cluster pair is
\begin{equation}
distance(C_\alpha, C_\beta)=\min_{{x\in{C_\alpha}}, y\in{C_\beta}} distance(x, y) \notag.
\end{equation}
We can define a distance threshold such that
the measured distance of a cluster pair should be smaller than
the threshold before merging.

\textbf{Spectral Clustering.}
Spectral clustering \cite{spectralClustring}
is a graph-based clustering algorithm.
The goal of this algorithm is to
find a partition of the graph that
minimizes the sum of weights on edges
between different clusters
while maximizing the sum of weights on edges within each cluster.
\begin{figure*}[!t]
\centering
\begin{tabular}{cc}
\includegraphics[width=0.43\linewidth]{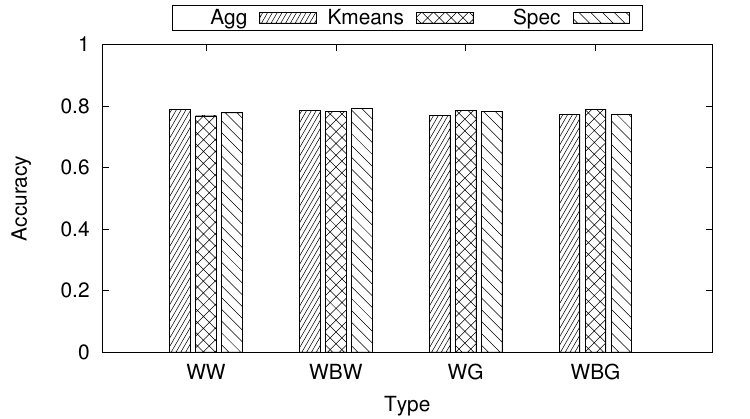}&
\includegraphics[width=0.43\linewidth]{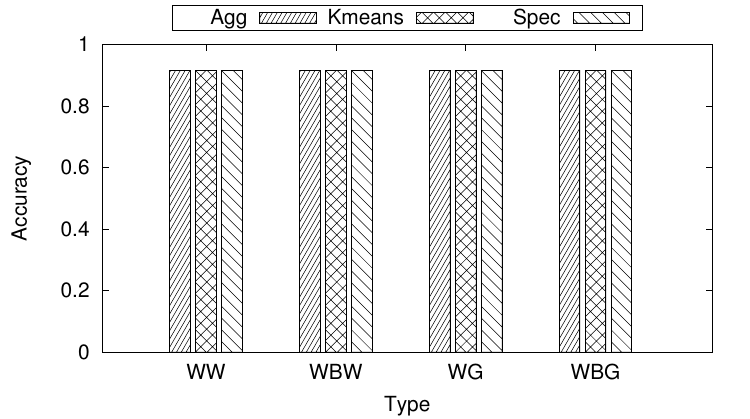}\\
(a)  & (b)\\
\end{tabular}
\caption{Testing accuracy of FedSAUC-trained models by (a) training set 1 and (b) training set 2.}
\label{Fig:testacc_all}
\end{figure*}

\subsection{Update Control}
In this component, we are going to
select active devices for the update
from the cluster set $C$ obtained from ``analysis of model similarity''.
If we always choose the same devices for the update,
it may consume the power and transmission bandwidth of these device
and also make the global model have a bias
to behaviors of these devices' owners.
Therefore, we propose a mechanism that selects devices randomly.
For each round, each device has a probability $\tau$ ($0 < \tau \leq 1$)
to be the representative of its cluster $C_k$.
They flip a coin to see if it is the representative or not.
Non-representative devices
may enter the power-saving mode in this round.
The probability $\tau$ is a system parameter
for controlling the number of representatives.
If we set $\tau$ to a higher value,
we will acquire more devices for the training process.
On the other hand, if we set $\tau$ to a lower value,
we may let more devices be inactive for saving their power.

\begin{figure*}[!t]
\centering
\begin{tabular}{cc}
\includegraphics[width=0.43\linewidth]{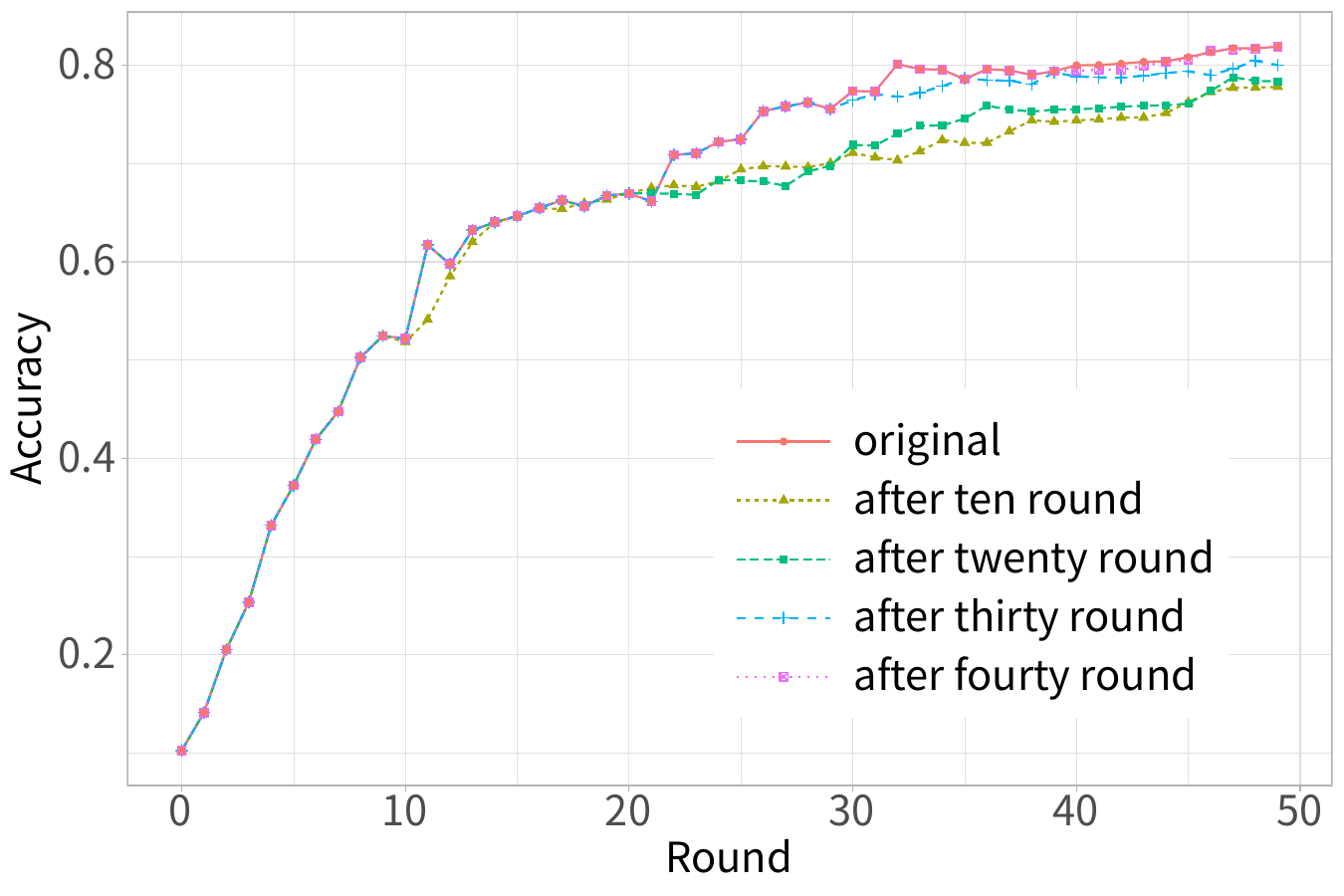}&
\includegraphics[width=0.43\linewidth]{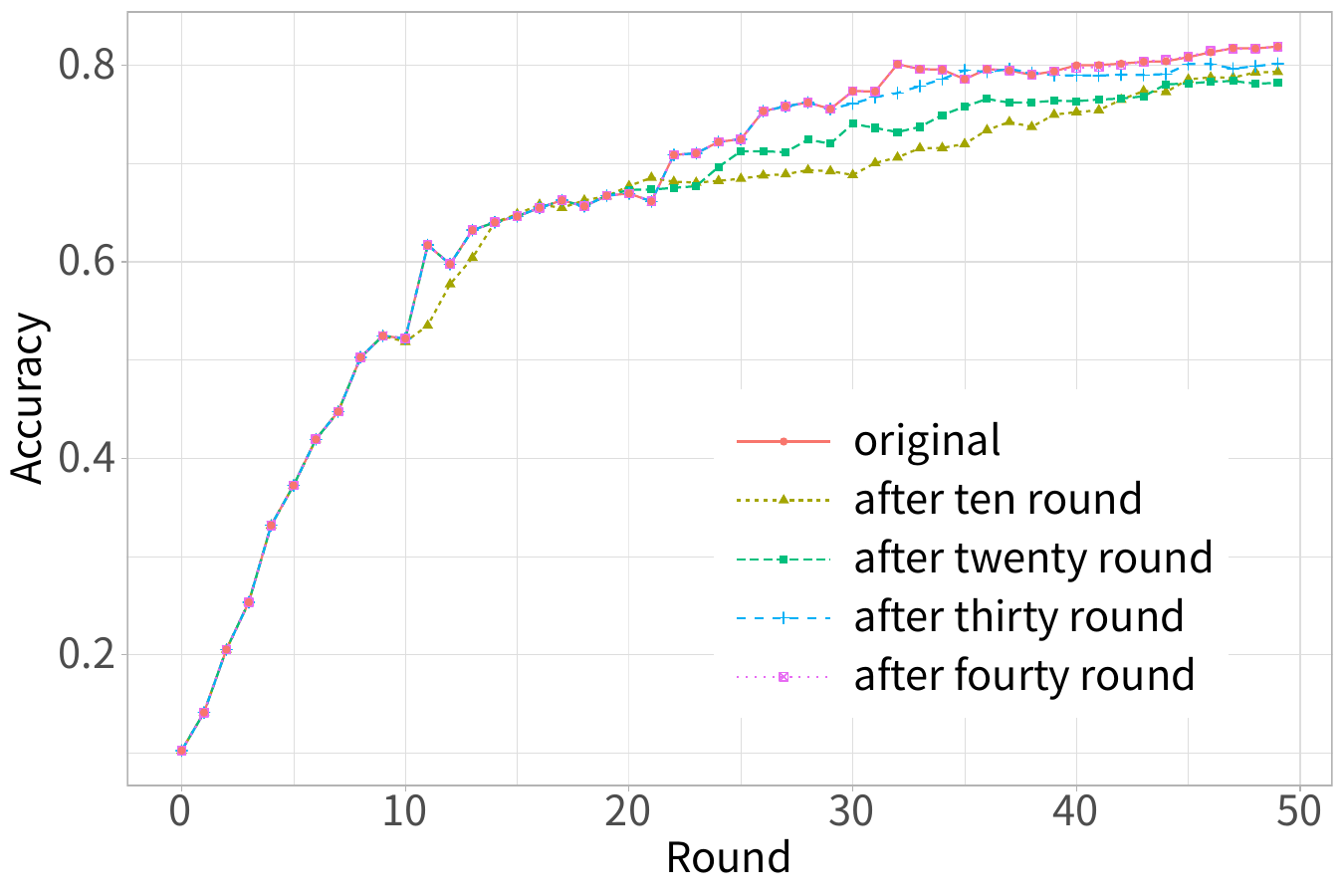}\\
(a) WW & (b) WBW\\
\includegraphics[width=0.43\linewidth]{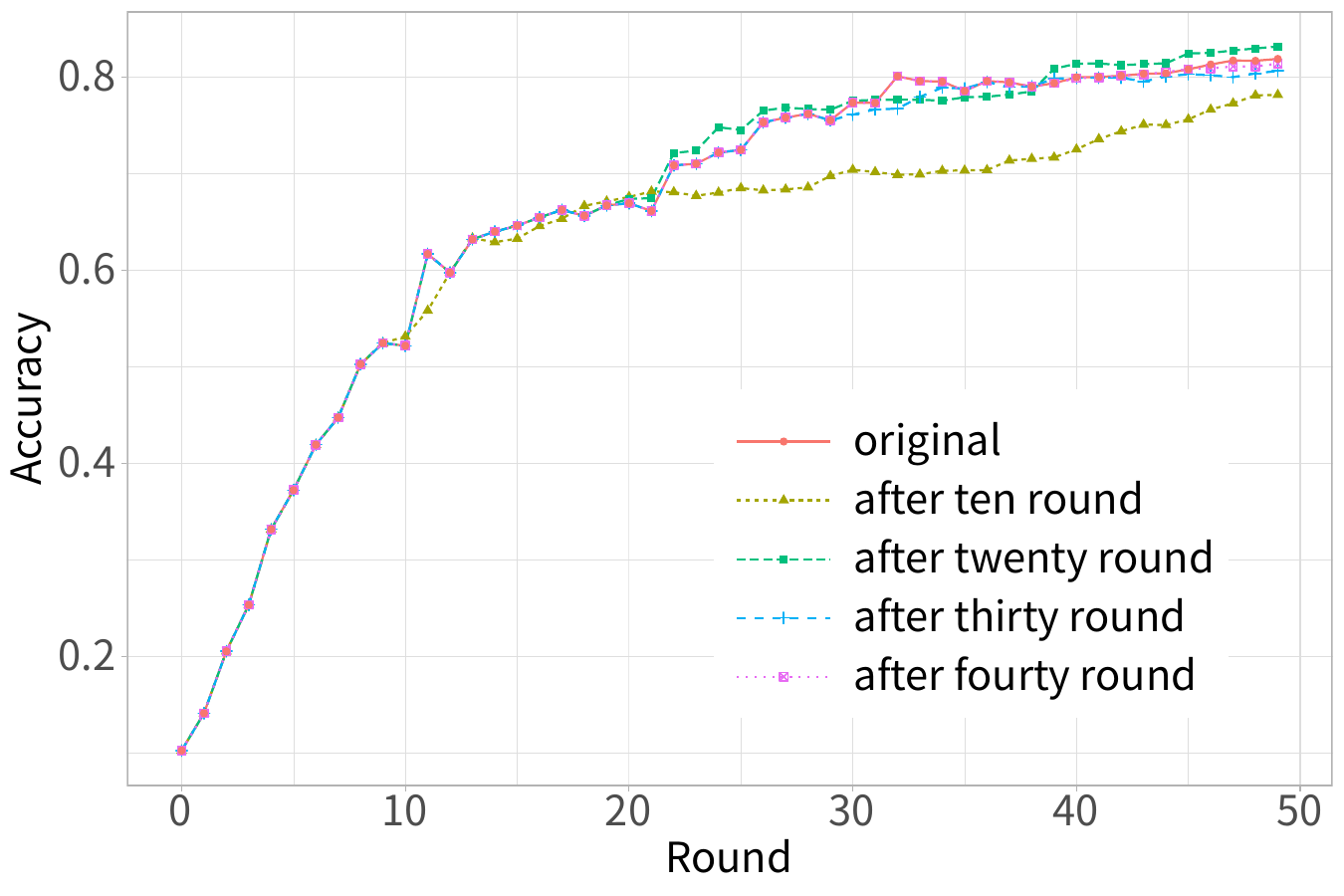}&
\includegraphics[width=0.43\linewidth]{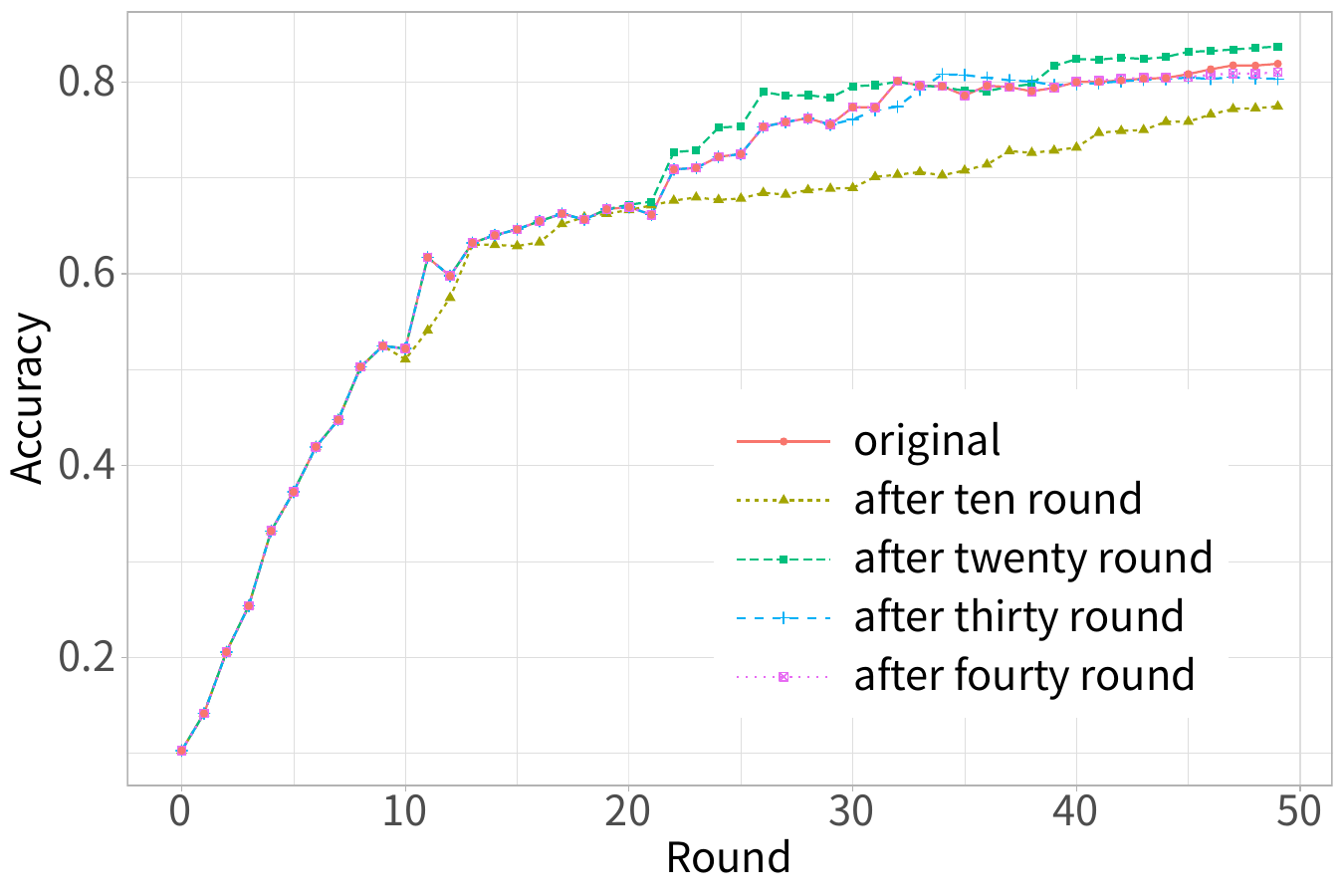}\\
(c) WG & (d) WBG\\
\end{tabular}
\caption{Testing accuracy of the model trained by dataset 1 when applying FedSAUC after different rounds.}
\label{Fig:testacc_raw_1}
\end{figure*}

\begin{figure*}[!t]
\centering
\begin{tabular}{cc}
\includegraphics[width=0.43\linewidth]{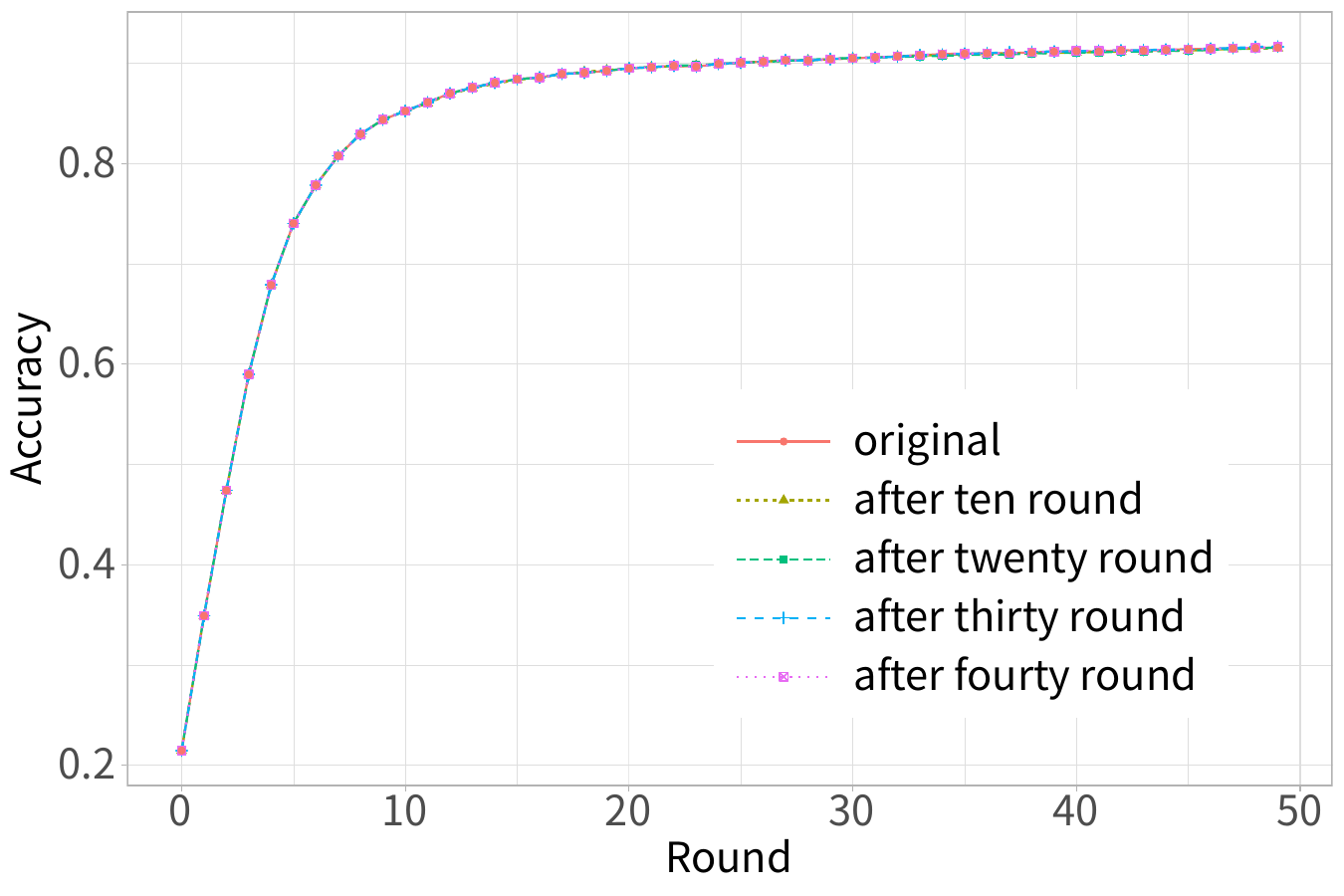}&
\includegraphics[width=0.43\linewidth]{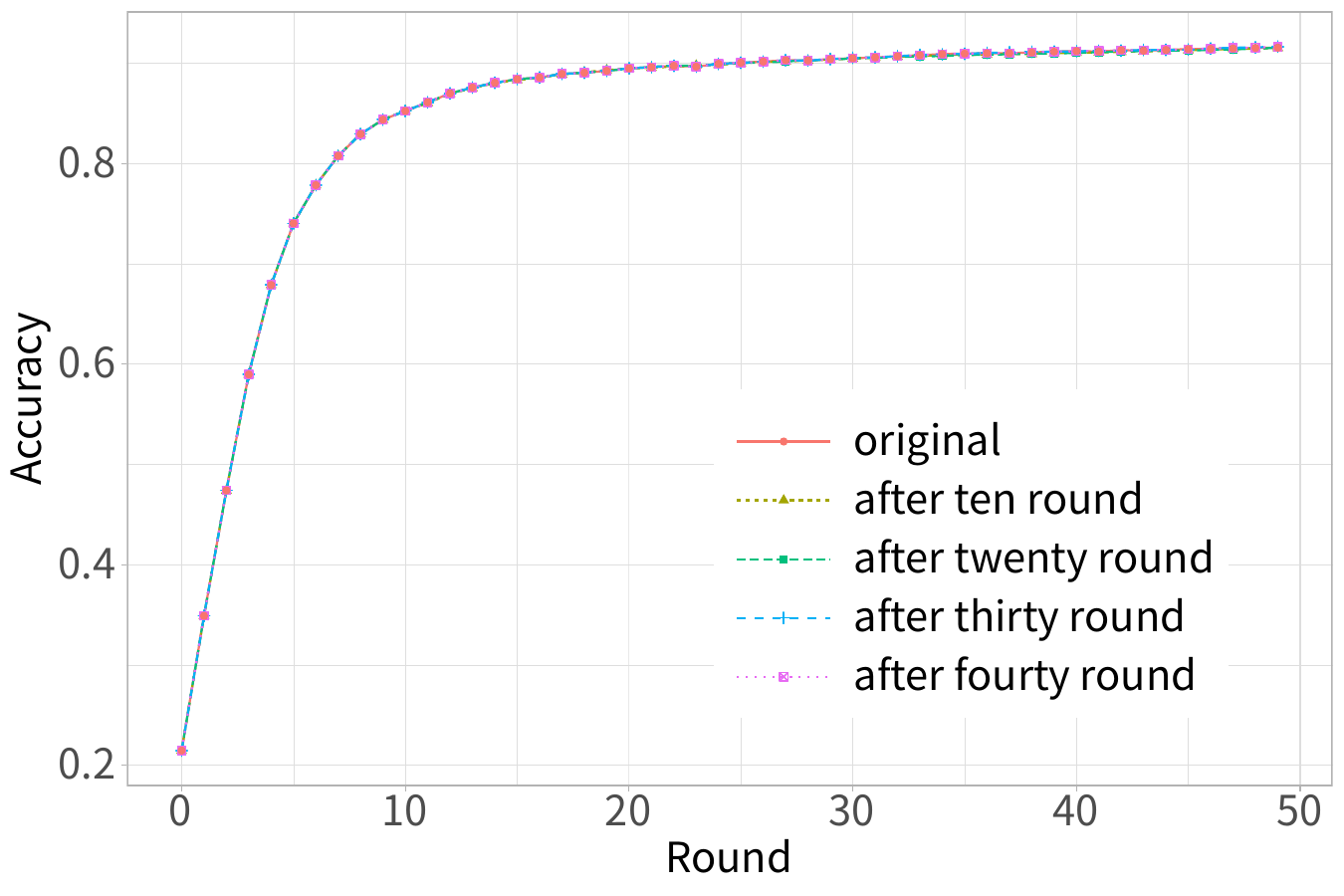}\\
(a) WW & (b) WBW\\
\includegraphics[width=0.43\linewidth]{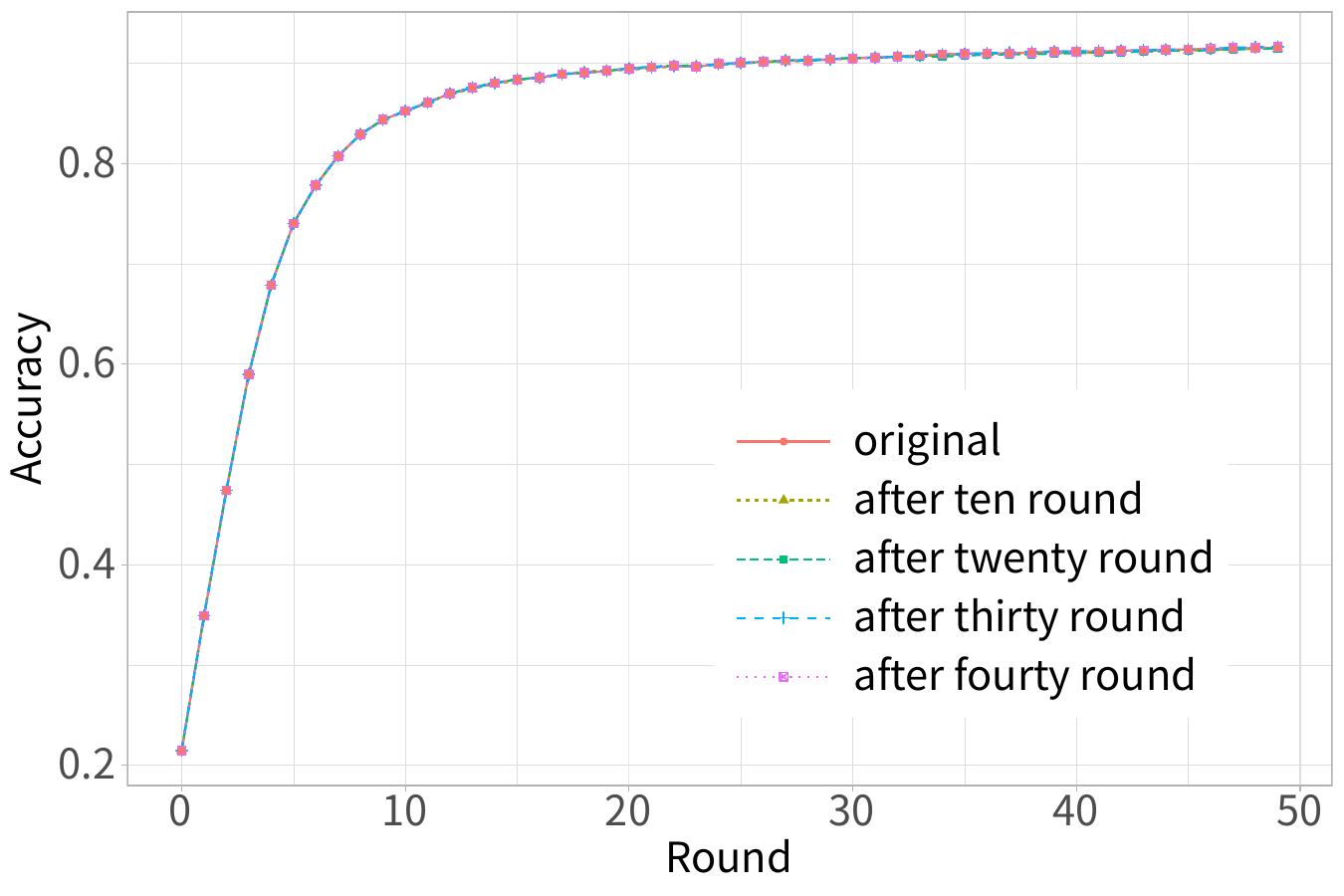}&
\includegraphics[width=0.43\linewidth]{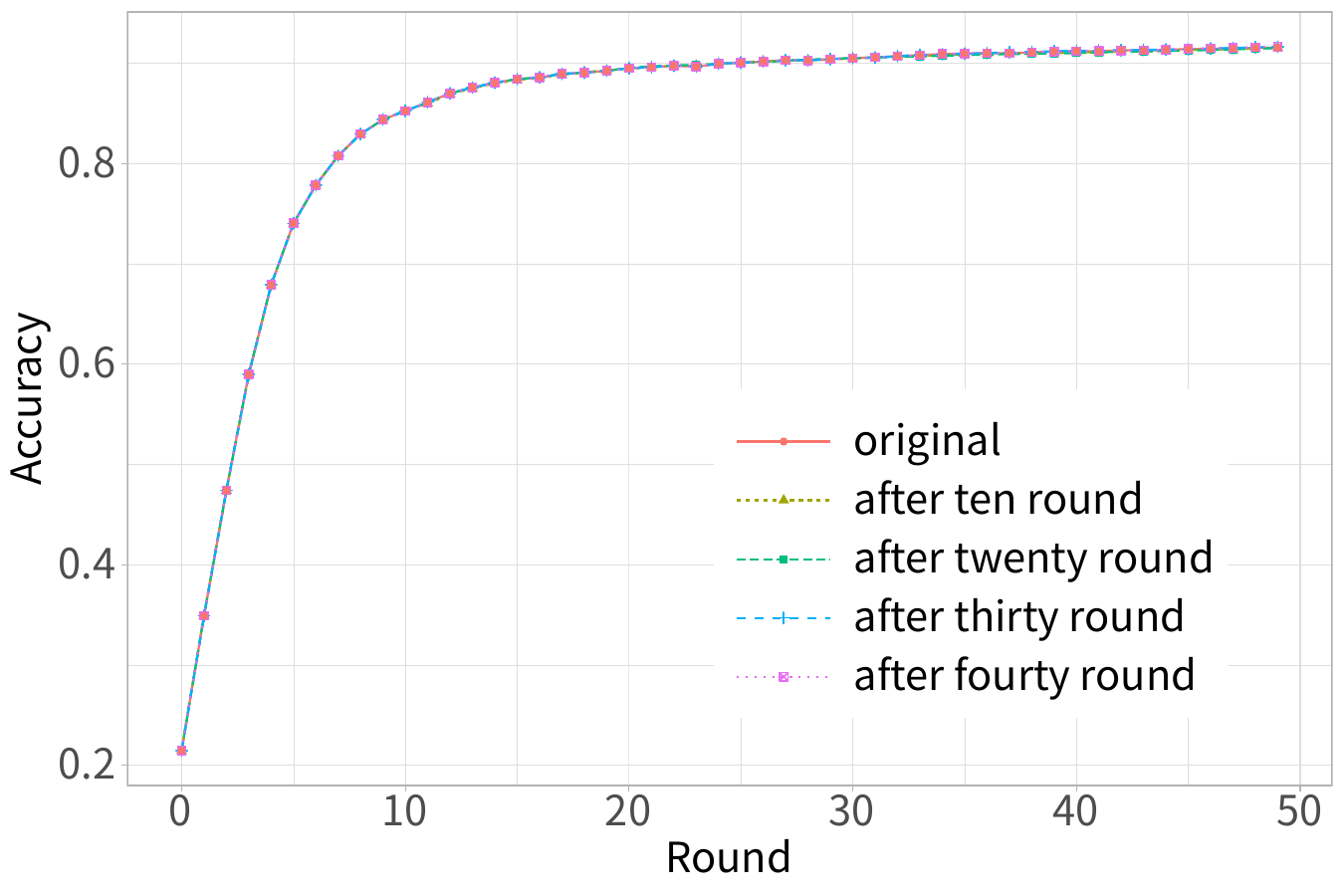}\\
(c) WG & (d) WBG\\
\end{tabular}
\caption{Testing accuracy of the model trained by dataset 2 when applying FedSAUC after different rounds.}
\label{Fig:testacc_our_1}
\end{figure*}

\section{Performance Evaluation}\label{Sec:Experimental Results}
\subsection{Experimental Environment Setup}
\textbf{Federated Learning Testbed on Edge Devices.}
To evaluate our proposed FedSAUC,
we setup a testbed of federated learning on edge devices
as shown in \Fig{Fig:Environment}.
There are five Raspberry Pi 4 ($v_0$ to $v_4$) and
six NVIDIA Jetson Nano ($v_5$ to $v_{10}$) in our testbed.
One Jetson Nano ($v_{10}$) serves as the server
of federated learning and
the rest of devices serve as clients.
All the devices are connected to a Gigabit Ethernet switch
via Ethernet cables.
We implemented FedSAUC on the top of the framework of FedML \cite{fedml},
which is an open research library and benchmark of federated learning.

\textbf{System Configurations.}
For evaluating the system performance,
we utilize linear regression model as the machine learning algorithm
to train by MNIST dataset \cite{mnist}, a handwriting digit dataset.
In order to differentiate the behaviors of devices' owners,
we partition MNIST dataset into two settings.
Each device trains and evaluates only on one or a pair of specific digit(s)
as shown in \Tab{tab:device}.
The \emph{total communication rounds} between the server and clients
is set to $50$ rounds,
\emph{epoch} of each round is set to $1$,
and the \textit{batch size} of each epoch is set to $32$.

\subsection{The Effectiveness of Model Similarity Analysis}
In \Sec{Sec:msim}, we use gradient updates $G$ for
the model similarity computation.
Here, we define $4$ types of updates for the performance comparison.
\begin{itemize}
  \item Use coefficients of the linear regression model \emph{(WW)} for the update.
  \item Use coefficients and also intercept (bias) of the linear regression model \emph{(WBW)} for the update.
  \item Use gradient updates of coefficients for the update \emph{(WG)}.
  \item Use gradient updates of coefficients and intercept for the update \emph{(WBG)}.
\end{itemize}
We implement these three clustering algorithms
by Scikit-Learn \cite{scikit-learn}.
\Tab{tab:clustering} shows the clustering performance
of these three algorithms
with respect to four different kinds of updates.
The accuracy in \Tab{tab:clustering} is computed by
the accumulated number of corrected grouping divided by
the accumulated number of groups.
As we can see,
the performance of spectral clustering is slightly
better than that of the other two methods.
In addition,
using gradient updates for the clustering
is better than using coefficients of linear regression.
The clustering performance of training set 2 is
much better than training set 1.
The reason is that
there are two digit types for each device group in training set 1
and they may mix the samples during the training process.
In some round, the devices in the same group
are training on totally different digits
but we label them a group
according to our device training configuration.
For training set 2,
there is only one digit type for each group.
Thus, it does not have the problem as training set 1.


\subsection{Communication Efficiency}
After applying update control,
some devices can skip the update transmission.
\Tab{tab:reduction} shows the overall reduction
of entire system when applying update control
from round $10$, $20$, $30$, and $40$.
In MNIST dataset,
there are images of $10$ digits and
the dimension of each image is $28 \times 28$.
Thus, we have $7840 = 28 \times 28 \times 10$ parameters (coefficients)
in one-dimensional linear regression.
If we set $\tau = 0.5$ in update control,
we can compute the reduction for each device
by $7840 \times (50 - r) \times 0.5$ on average
where $r$ is the round number when applying update control.

\subsection{Training Performance}
In this section, we investigate the performance comparison of model accuracy
when applying FedSAUC on different update settings and rounds.
\Fig{Fig:testacc_all} shows that the testing accuracy
of the models trained by applying FedSAUC after round 10
using these three clustering algorithms.
The results show that the testing accuracy
is very close for each training set under different clustering methods.
In \Fig{Fig:testacc_raw_1}, we vary the time of applying FedSAUC
to see the influences of FedSAUC on the model testing accuracy.
We can see that if we apply FedSAUC after round 10,
the testing accuracy decreases about $10\%$
since we use fewer data to train the model.
Nevertheless, the test accuracy is approaching to
the accuracy of the model without applying FedSAUC at round $50$.
Also, the update settings do not greatly affect the model accuracy.
In \Fig{Fig:testacc_our_1},
the accuracies under different settings are very close.
We think that it is because
only one digit label for each group in the training set 2
make the classification easier with fewer data.

\section{Conclusions}\label{Sec:conc}
In this work,
we presented an update control mechanism, FedSAUC, for federated learning
to save the energy consumption and bandwidth of edge devices.
We considered the model similarity to
keep the data diversity for the learning process.
In our experiment, we showed that
the model accuracy may decrease because of
our update control.
But, it will be close to the accuracy of regular federated learning
in the long run.
Therefore,
we can utilize the update control for
saving the energy and network resource on edge devices
at the expense of model accuracy.
We also validated the prototyping of federated learning
with our method on edge devices.

\section*{Acknowledgement}
\thanks{
    Y.-A. Chen's research
    is co-sponsored by
    MOST 107-2218-E-155-007-MY3, Taiwan, 
    MOST 110-2221-E-155-022-MY3, Taiwan, 
    and Innovation Center for Big Data and Digital Convergence, Yuan Ze University, Taiwan.
}

\bibliographystyle{IEEEtran}
\bibliography{bibfile}
\end{document}

%% file: macros.tex
\newcommand{\blist}{\begin{list}\setlength{\topsep 0pt \parsep 0pt}}
\newcommand{\elist}{\end{list}}
\newtheorem{thm}{Theorem}
\newcommand{\bthm}{\begin{thm}\begin{textit}}
\newcommand{\ethm}{\end{textit}\end{thm}}
\newtheorem{dfn}{Definition}
\newcommand{\bdfn}{\begin{dfn}\begin{textit}}
\newcommand{\edfn}{\end{textit}\end{dfn}}
\newtheorem{obn}{Observation}
\newcommand{\bobn}{\begin{obn}\begin{textit}}
\newcommand{\eobn}{\end{textit}\end{obn}}

\newtheorem{lema}{Lemma}
\newcommand{\bdla}{\begin{lema}\begin{textit}}
\newcommand{\edla}{\end{textit}\end{lema}}
\newtheorem{pro}{Property}
\newcommand{\bpro}{\begin{pro}\begin{textit}}
\newcommand{\epro}{\end{textit}\end{pro}}

\newcounter{algnum1}

\newcounter{algnum2}

\newcounter{algnum3}

\newcommand\bit{\begin{itemize}}
\newcommand\eit{\end{itemize}}

\newcommand{\Fig}[1]{Fig.~\ref{#1}}

\newcommand{\Sec}[1]{Section~\ref{#1}}
\newcommand{\Tab}[1]{Table~\ref{#1}}

\newcommand{\balgo}{
   \begin{description}
   \parskip=0pt
   \topsep=0pt
}
\newcommand{\ealgo}{
   \end{description}
}

%% file: main.bbl
\begin{thebibliography}{10}
\providecommand{\url}[1]{#1}
\csname url@samestyle\endcsname
\providecommand{\newblock}{\relax}
\providecommand{\bibinfo}[2]{#2}
\providecommand{\BIBentrySTDinterwordspacing}{\spaceskip=0pt\relax}
\providecommand{\BIBentryALTinterwordstretchfactor}{4}
\providecommand{\BIBentryALTinterwordspacing}{\spaceskip=\fontdimen2\font plus
\BIBentryALTinterwordstretchfactor\fontdimen3\font minus
  \fontdimen4\font\relax}
\providecommand{\BIBforeignlanguage}[2]{{%
\expandafter\ifx\csname l@#1\endcsname\relax
\typeout{** WARNING: IEEEtran.bst: No hyphenation pattern has been}%
\typeout{** loaded for the language `#1'. Using the pattern for}%
\typeout{** the default language instead.}%
\else
\language=\csname l@#1\endcsname
\fi
#2}}
\providecommand{\BIBdecl}{\relax}
\BIBdecl

\bibitem{mcmahan2017communication}
B.~McMahan, E.~Moore, D.~Ramage, S.~Hampson, and B.~A. y~Arcas,
  ``Communication-efficient learning of deep networks from decentralized
  data,'' in \emph{Proc. of International Conference on Artificial Intelligence
  and Statistics (AISTATS)}, 2017.

\bibitem{fl_chanllenges}
T.~Li, A.~K. Sahu, A.~Talwalkar, and V.~Smith, ``Federated learning:
  Challenges, methods, and future directions,'' \emph{IEEE Signal Processing
  Magazine}, vol.~37, no.~3, pp. 50--60, 2020.

\bibitem{hudecentralized}
C.~Hu, J.~Jiang, and Z.~Wang, ``Decentralized federated learning: a segmented
  gossip approach,'' \emph{arXiv preprint arXiv:1908.07782}, 2019.

\bibitem{decentralized}
Z.~Tang, S.~Shi, and X.~Chu, ``Communication-efficient decentralized learning
  with sparsification and adaptive peer selection,'' in \emph{Proc. of
  International Conference on Distributed Computing Systems (ICDCS)}, 2020.

\bibitem{semiasyn}
W.~Wu, L.~He, W.~Lin, R.~Mao, C.~Maple, and S.~A. Jarvis, ``Safa: a
  semi-asynchronous protocol for fast federated learning with low overhead,''
  \emph{IEEE Transactions on Computers}, vol.~70, no.~1, pp. 655--668, 2021.

\bibitem{dgc}
Y.~Lin, S.~Han, H.~Mao, Y.~Wang, and W.~Dally, ``{Deep Gradient Compression:
  Reducing the Communication Bandwidth for Distributed Training},'' in
  \emph{Proc. of International Conference on Learning Representations (ICLR)},
  2018.

\bibitem{expanding}
S.~Caldas, J.~Kone{\v{c}}ny, H.~B. McMahan, and A.~Talwalkar, ``Expanding the
  reach of federated learning by reducing client resource requirements,''
  \emph{arXiv preprint arXiv:1812.07210}, 2018.

\bibitem{qgmu}
M.~M. Amiri, D.~Gunduz, S.~R. Kulkarni, and H.~V. Poor, ``Federated learning
  with quantized global model updates,'' \emph{arXiv preprint
  arXiv:2006.10672}, 2020.

\bibitem{ds}
S.~Ji, W.~Jiang, A.~Walid, and X.~Li, ``Dynamic sampling and selective masking
  for communication-efficient federated learning,'' \emph{arXiv preprint
  arXiv:2003.09603}, 2020.

\bibitem{untargeted}
R.~A. Mallah, D.~Lopez, and B.~Farooq, ``Untargeted poisoning attack detection
  in federated learning via behavior attestation,'' \emph{arXiv preprint
  arXiv:2101.10904}, 2021.

\bibitem{gan}
J.~Zhang, B.~Chen, X.~Cheng, H.~T.~T. Binh, and S.~Yu, ``Poisongan: Generative
  poisoning attacks against federated learning in edge computing systems,''
  \emph{IEEE Internet of Things Journal}, vol.~8, no.~5, pp. 3310--3322, 2021.

\bibitem{kmeans}
T.~Kanungo, D.~M. Mount, N.~S. Netanyahu, C.~D. Piatko, R.~Silverman, and A.~Y.
  Wu, ``An efficient k-means clustering algorithm: Analysis and
  implementation,'' \emph{IEEE Transactions on Pattern Analysis and Machine
  Intelligence}, vol.~24, no.~7, pp. 881--892, 2002.

\bibitem{agg}
W.~H. Day and H.~Edelsbrunner, ``Efficient algorithms for agglomerative
  hierarchical clustering methods,'' \emph{Journal of classification}, vol.~1,
  no.~1, pp. 7--24, 1984.

\bibitem{spectralClustring}
U.~Von~Luxburg, ``A tutorial on spectral clustering,'' \emph{Statistics and
  computing}, vol.~17, no.~4, pp. 395--416, 2007.

\bibitem{fedml}
C.~He, S.~Li, J.~So, M.~Zhang, H.~Wang, X.~Wang, P.~Vepakomma, A.~Singh,
  H.~Qiu, L.~Shen, P.~Zhao, Y.~Kang, Y.~Liu, R.~Raskar, Q.~Yang, M.~Annavaram,
  and S.~Avestimehr, ``Fedml: A research library and benchmark for federated
  machine learning,'' \emph{arXiv preprint arXiv:2007.13518}, 2020.

\bibitem{mnist}
``{THE MNIST DATABASE} of handwritten digits,''
  \url{http://yann.lecun.com/exdb/mnist/}.

\bibitem{scikit-learn}
F.~Pedregosa, G.~Varoquaux, A.~Gramfort, V.~Michel, B.~Thirion, O.~Grisel,
  M.~Blondel, P.~Prettenhofer, R.~Weiss, V.~Dubourg, J.~Vanderplas, A.~Passos,
  D.~Cournapeau, M.~Brucher, M.~Perrot, and E.~Duchesnay, ``Scikit-learn:
  Machine learning in {P}ython,'' \emph{Journal of Machine Learning Research},
  vol.~12, pp. 2825--2830, 2011.

\end{thebibliography}
